\documentclass[unnumsec,webpdf,modern,large,numbered]{oup-authoring-template}

\usepackage{amsmath}
\usepackage{booktabs}
\usepackage{tabularx}
\usepackage{array}
\makeatletter
\def\ps@curemanuscript{%
  \def\@oddhead{\hbox to\textwidth{\footnotesize\sffamily
    From annotation to reasoning\hfil Perspective}}%
  \let\@evenhead\@oddhead
  \def\@oddfoot{\hfil\footnotesize\sffamily\thepage\hfil}%
  \let\@evenfoot\@oddfoot}
\let\ps@opening\ps@curemanuscript
\let\ps@headings\ps@curemanuscript
\makeatother

\newcolumntype{Y}{>{\raggedright\arraybackslash}X}
\hypersetup{
  pdftitle={From annotation to reasoning: Culture in language models},
  pdfauthor={Daniel Hershcovich; Alexander Conroy; Jens Bjerring-Hansen},
  pdfsubject={Perspective},
  pdfkeywords={cultural reasoning, language models, literary interpretation, evaluation, responsible AI}
}

\begin{document}

\journaltitle{}
\DOI{}
\copyrightyear{}
\pubyear{}
\vol{}
\issue{}
\access{}
\appnotes{Perspective}
\firstpage{1}

\title[From annotation to reasoning]{From annotation to reasoning: Culture in language models}

\author[1,$\ast$]{Daniel Hershcovich}
\author[2]{Alexander Conroy}
\author[2]{Jens Bjerring-Hansen}
\address[1]{\orgdiv{Department of Computer Science}, \orgname{University of Copenhagen}, \orgaddress{\street{Vermundsgade 5}, \postcode{2100}, \state{Copenhagen}, \country{Denmark}}}
\address[2]{\orgdiv{Department of Nordic Studies and Linguistics}, \orgname{University of Copenhagen}, \orgaddress{\state{Copenhagen}, \country{Denmark}}}
\corresp[$\ast$]{Corresponding author: Daniel Hershcovich, \href{mailto:dh@di.ku.dk}{dh@di.ku.dk}. ORCID: \href{https://orcid.org/0000-0002-3966-8708}{0000-0002-3966-8708}.}

\abstract{How should we evaluate language models when more than one interpretation can be right? Cultural benchmarks often test factual knowledge, agreement with survey responses, or recognition of a predefined meaning. These tasks leave open whether a model can explain how a cultural reference works in a particular text, support a reading with evidence, or revise it after criticism. This is a question of interpretive depth, complementary to the breadth of cultural coverage. We argue that literary interpretation offers a useful setting for studying these capabilities. We focus on cultural referencing and reuse: how texts invoke, repeat, and transform earlier expressions across historical and linguistic contexts. Our central claim is that literary scholars can disagree about an interpretation while recognizing the quality of its support. We propose linking evidence-centered benchmarks, evaluation that preserves scholarly disagreement, and model-development experiments on literary data, contextual resources, and scholarly feedback. Danish literature provides a concrete starting point, with implications for other languages and domains. The aim is to develop alternative evaluation strategies that go beyond conventional benchmark metrics and guide model development toward cultural robustness in AI systems.}

\keywords{cultural reasoning, language models, literary interpretation, evaluation, responsible AI}

\maketitle
\pagestyle{curemanuscript}
\raggedbottom

\section{What would count as cultural reasoning?}

How good are large language models (LLMs) at cultural reasoning? Can we make them better? What does \lq\lq better'' even mean in this case? These questions become difficult when we ask a model to interpret a text. An allusion, for example, draws on an earlier source and acquires meaning through its use in a new context. Evaluating a model's account requires attention to both the connection and its significance for the passage.

Cultural evaluation has devoted considerable attention to \emph{breadth}: the range of languages, communities, and cultural knowledge represented in models and benchmarks \citep{hershcovich2022challenges,liu2025culturally}. Expanding this coverage helps examine whose knowledge and values models represent. In their survey, \citet{chakraborty2026datasets} identify a tension between scalable dataset construction and the depth and conceptual precision of cultural representation. Cultural knowledge, population-level alignment, and interpretation each pose distinct evaluation questions. Survey distributions can reveal whether a model systematically favors certain responses \citep{tao2024cultural}. Interpreting a proverb used ironically requires an account of the speaker, context, and relation between the familiar expression and its present use. In each case, an evaluation must establish how the observed response supports a claim about the capability being assessed \citep{tedeschi2023meaning,jacobs2021measurement}.

Our focus is \emph{interpretive depth}: how well a model explains the significance of an expression in context, supports its reading with evidence, considers plausible alternatives, and responds to criticism. Breadth and depth are complementary aims, requiring evidence about both the range of contexts covered and the quality of interpretation within them. We evaluate the reasons a model offers and the support for those reasons. Generated explanations can diverge from the internal computation that produced an answer \citep{jacovi2020faithfully,turpin2023unfaithful}; their evidential quality requires scrutiny even when the prose is fluent.

Three questions follow: How can we construct tasks that require culturally grounded interpretation? How can we evaluate responses while preserving ambiguity and disagreement? How can we adapt models without reinforcing stereotypes or reducing several defensible readings to one? These questions connect benchmark design, evaluation, and model development. They also motivate CuRe, our collaboration between natural language processing and literary studies. Danish literature provides the initial setting for this work, with implications for evaluation across languages and domains.

The distinction we need is straightforward: agreeing with an interpretation and judging it to be well supported are different acts. Two scholars may prefer different readings while recognizing that both account for the text, or agree that a third interpretation is inconsistent with the textual evidence. This scenario is familiar in literary practice and cultural interpretation \citep{eco1990limits}. Establishing its reliability as a basis for model evaluation requires empirical testing.

\section{Culture through references and reuse}

Literature is an active medium of cultural expression. Texts circulate expressions, revisit histories, and question the conventions on which their readers rely \citep{williams1983writing}. Interpreting them requires attention to perspective and ambiguity: two texts may use the same phrase, but in different contexts, which effectively alters its referentiality and meaning. This cultural condition makes literature a useful setting for studying culturally grounded reasoning.

Our starting point is culture as learned, shared, and contested patterns of meaning and practice. Population regularities and perceived shared norms can be informative \citep{gelfand2011differences,chiu2010intersubjective}, while people participate in overlapping cultural traditions that change through exchanges among communities \citep{morris2015polycultural}. A claim about a population consequently needs different evidence from a claim about an individual speaker or a literary work. Cultural NLP has good reason to question national labels as sufficient descriptions of either \citep{zhou2025culture}.

We approach cultural specificity through relationships between texts, drawing on work on cultural recurrence and computational literary studies \citep{resina2019repetition,piper2018enumerations}. Two phenomena are particularly useful. \emph{Cultural referencing} includes intertextual signals ranging from direct quotations to subtle allusions across national and historical boundaries. \emph{Cultural reuse} includes the repetition and transformation of culturally situated expressions, from elaborate formulations to worn-out clich\'es. Their recurrence allows us to trace how cultural knowledge and expression develop as texts draw on, alter, and contest earlier formulations.

In Karen Blixen's short story ``The Supper at Elsinore'', the elderly Madam B\ae k reaches Copenhagen after a difficult winter journey. The narrator compares her arrival to entering the ``new Jerusalem'', then describes ordinary city streets without the biblical gold and precious stones \citep{blixen1934seven}. An interpretive task could ask what this allusion contributes to the portrayal of Madam B\ae k. Revelation 21:2 and 18--21 helps identify the reference; the story's details determine how it functions. One reading could connect the heavenly imagery to her declining health and approaching death. Another could emphasize the gentle comedy of imagining her inspecting heaven with the same practical judgment she brings to Copenhagen. Both would need to account for her acceptance of the city and for the narrator's role in making the comparison. These emphases can coexist; their relative plausibility depends on how well the account relates the biblical imagery to the story's narrative and historical context.

Retrieval could locate the biblical imagery and passages about Madam B\ae k's health and journey. The model then needs to explain how these details support its reading and where another emphasis remains plausible. Literary scholars can examine these connections and probe how the model responds to further textual evidence or scholarly commentary.

Danish literature offers a historically, linguistically, and culturally varied repertoire in which to investigate these relationships. Existing corpora of historical Danish and Norwegian fiction and contemporary writing situated within global literary trends provide resources for this work \citep{bjerringhansen2022mending,piper2025miniworldlit}. Following references and reuse across such material allows us to study the evolving patterns that characterize Danish literature and its exchanges with other literary traditions. Historical survival, digitization, and canon formation shape which texts are available; documenting these choices helps establish the scope of claims about what a model has learned.

\section{Benchmarks that require interpretation}

An interpretive benchmark should begin with a question grounded in literary scholarship. \lq\lq Why is this passage ironic?'' or \lq\lq What changes when this expression is reused here?'' can require more than selecting a predefined meaning. Drawing on educational assessment, \citet{liu2024ecbd} argue for specifying the capability being assessed, the task that could elicit it, and the evidence needed to justify the assessment. \citet{sui2025kristeva} apply close-reading tasks to the evaluation of literary reasoning. Open-ended assignments can extend this work by making the grounds for an answer available for assessment.

We propose assignments developed with literary scholars and scholarly resources. Categorical annotations can record where an allusion or ambiguous expression occurs. Multiple human-written responses can illustrate different ways of answering the question, while leaving room for further readings. A model's unanticipated interpretation must also be evaluated against the text and the task.

Table~\ref{tab:tasks} illustrates how the same material can support several tasks. Together, they examine the identification of a reference, its contribution to a passage, and the revision of an account in response to criticism. Each task supplies evidence about a particular aspect of interpretation.

Tasks should distinguish plausible performance from shortcuts. A familiar canonical work may invite a model to reproduce commentary encountered during training. Less familiar material, documented source access, and held-out works can help test transfer~\citep{conroy2026cultural}. Scholars can compare how a model accounts for existing uses of a recurring expression by different speakers or in different works. These comparisons preserve the source texts and draw on documented differences in their narrative and historical contexts.

Long-context evaluation matters for the same reason. A passage may support one interpretation in its immediate context, only for a later scene to challenge that reading. Comparing passage-only, full-text, and retrieval-assisted conditions can identify what additional context enables a model to do. These comparisons should also test whether models overlook relevant material or use a retrieved source in a misfitting temporal or cultural context.

Coverage across languages, periods, and genres should be reported alongside the quality of evidence use and revision within each setting. This would allow us to test whether broader cultural coverage also improves interpretive depth, and where each requires further development.

\section{Judging quality without requiring agreement}

Evaluation should draw on literary scholars with relevant linguistic and historical expertise and familiarity with the works under assessment. Work on human label variation shows why majority aggregation can discard meaningful differences \citep{plank2022problem,uma2021learning,davani2022dealing}. Retaining individual judgments and using probabilistic soft labels where response categories permit them can preserve these differences. Multiple scholarly reference interpretations can guide assessment of open-ended responses while leaving room for further well-supported readings.

Describing the range of positions is one part of evaluation. Resources that organize social norms, recurring model justifications, and contextualized cultural descriptions make this variation available for study \citep{forbes2020social,wright2024tropes,shi2024culturebank}. The quality of support within that range raises a further question: a frequent position can be poorly argued, and an uncommon reading can be carefully supported. Work on pluralistic alignment emphasizes representing diverse positions \citep{sorensen2024roadmap}. Interpretive evaluation adds an assessment of what makes a particular position defensible.

We propose separating three judgments. First, does a response contain identifiable errors, such as invented evidence or a misattributed statement? Second, how well does it connect accurate evidence and relevant context to the interpretation, including its treatment of alternatives? Third, how plausible do the expert evaluators judge the different interpretations to be? Plausibility admits degrees, and several accounts may receive substantial support. The distinction builds on argument-quality assessment and the relation between a claim, its evidence, and its warrant---the reason the evidence supports that claim \citep{toulmin1958uses,wachsmuth2017argumentation}. It allows scholars to assess the quality of an account while differing in the weight they give it.

The evaluation record can be simple: the question and available materials, the proposed reading, cited passages, contextual assumptions, the connection between them, and what might change the account. It should make clear whether a claim concerns a particular passage, a genre, or an epoch. Experts can then identify where their judgments diverge. Evaluators need access to this information across different forms of response; a rigid template risks rewarding compliance and verbosity.

The protocol should retain each expert's assessments and explanations. Where a task specifies mutually exclusive alternatives, experts can assign a probability distribution reflecting their relative plausibility; uncertainty within an assessment and disagreement between experts should remain distinguishable. Interpretations that can coexist, such as the emphases on mortality and humor in the Blixen example, require separate graded assessments. Reports should describe both the distribution of judgments and the reasons for them. Comparing model responses before and after scholarly feedback can then show whether a revision improves evidential support, addresses an objection, or accommodates a further plausible account.

Professional judgments are shaped by interpretive communities \citep{fish1980text}, whose conventions can also exclude voices and naturalize assumptions. The proposed evaluation relies on specialists working from identified editions and relevant scholarly resources, with their expertise and familiarity with the works documented. Scholarly consensus can provide a provisional point of reference, while competing accounts and their grounds remain available for scrutiny. Criteria can be challenged along with model outputs. Factual accuracy and relevance constrain the range of defensible interpretations within this process.

Comparisons across languages raise a further problem: translating questions or rating terms can change what they measure \citep{chen2008chopsticks}. Rubrics need local adaptation and evidence of comparability before scores are pooled. Separate profiles can preserve differences between evaluative traditions where pooling would obscure them. Automated judges need equivalent validation against human assessments and diagnostic cases, including checks for shared model preferences that may inflate agreement.

\section{What should models learn from literature?}

The contribution of literary training data depends on which capabilities we evaluate. In a Norwegian training-data study, \citet{delarosa2025impact} found that adding fiction reduced aggregate performance on the chosen benchmarks while improving textual variation and readability. They call for evaluations of capabilities such as plot understanding and creative writing. The result leaves fiction's contribution to cultural interpretation unresolved, and illustrates why the choice of evaluation matters for decisions about training data.

Training-data comparisons should ask how historical and contemporary fiction, general-language material, and scholarly commentary contribute to interpretive performance. Controlling the amount of training and documenting data sources would help isolate the contribution of corpus composition. Testing both familiar expressions and new uses of them would help distinguish memorized commentary from the transfer of interpretive capabilities.

Interpretation also depends on access to relevant context. Providing scholarly commentary, literary history, or analysis through retrieval differs from including it in pretraining or fine-tuning. Retrieval offers traceability and can supply material absent from training, but models still need to judge its relevance. Experiments should compare the same tasks with and without contextual resources and assess the relevance of the retrieved material to each task. This connects model development to an evaluation question: when does access to another source improve the analysis, and when does it merely make a weak claim look documented?

Scholarly feedback raises a further question: what should the model learn from an evaluator's judgment? Fine-tuning, preference learning, and reinforcement learning offer ways to investigate this. A preference for a whole essay gives a model little information about whether its evidence, contextual assumption, or inference was at fault. More localized feedback may be more useful, but it is also costly. Comparing the granularity of feedback and testing revision on held-out works would show whether that cost produces better interpretation. Training data should include several defensible accounts, with feedback explaining their respective strengths and weaknesses.

A model can learn the outward form of scholarly caution while continuing to make unsupported claims. It can also become so compliant that it abandons a well-grounded reading whenever challenged. Successful revision must distinguish a reason to change from a request to agree. Expert evaluation should therefore examine whether the model can justify retaining a well-supported claim as well as revising a weak one, including on works from other periods or genres. The range of these tests determines how broadly improvements can be expected to generalize.

These experiments ask which data, access to context, and learning signals improve the use of evidence while retaining interpretive alternatives. By cultural robustness we mean a model's ability to sustain contextually grounded reasoning across variations in language, period, genre, and interpretive standpoint. Evaluation should expose when adaptation reinforces a stereotype, treats a situated expression as a timeless cultural fact, or suppresses a well-supported alternative. Establishing how these effects vary across models and tasks would give developers a basis for choosing methods of cultural adaptation.

\section{Expert scrutiny of cultural reasoning}

The intended use is an evaluation framework in which expert readers probe a model's capacity to recognize connections, locate sources, and reason about cultural alternatives. In this interaction, literary scholars can challenge a claim, question an assumption, or ask what evidence would distinguish two readings. Its value lies in exposing the model's interpretive limits and testing whether it can sustain a culturally grounded argument when challenged by an expert.

Teaching provides an example of how model outputs can prompt examination of interpretive assumptions. In \citet{bilstrup2026sentiment}, students compared their sentiment annotations with model predictions and discussed the reasons for disagreement. A classification task became an occasion to examine narrative perspective and context. For model evaluation, such disagreements motivate questions about which features of a text support the output and how the system accounts for them.

Book summarization raises similar questions~\citep{conroy2026cultural}. Multiple summaries can be valid, yet factual and contextual errors remain assessable. \citet{allaith2026novelsum} evaluated summaries of 19th century Scandinavian novels using general and narrative-specific criteria and found automated assessment less dependable for factuality and setting-related criteria. A summary is also a selection: which events or characters does it foreground, for whom, and why? Expert evaluation can examine those choices against the passages and narrative contexts on which they depend.

The selections made in summaries and interpretations can also shape which cultural accounts circulate. A related concern emerges from a controlled writing study with participants in India and the United States: \citet{agarwal2025suggestions} found that AI suggestions shifted Indian participants' writing toward Western styles. For interpretation, this raises a question that needs separate testing: do models repeatedly privilege familiar or dominant readings? Expert evaluation can examine the alternatives a system offers, the cultural assumptions it invokes, and the interpretations it overlooks.

The practical requirement is to keep sources, disagreement, and correction accessible. Evaluation should track which interpretations a model represents, how it supports them, and whether scholarly criticism leads to substantive revision. These assessments can guide choices about training data, retrieval, and feedback for culturally responsible model development.

\section{A broader role for interpretive evaluation}

The separation between a conclusion and its support also matters in other fields. Visual cultural interpretation asks how symbols are reused in particular objects and settings \citep{yadav2025cultural}. Law requires reasons that can be examined against sources, doctrine, and institutional procedures. In a small study of European Court of Human Rights cases, \citet{raina2026legal} found that more comprehensive model analysis did not improve outcome accuracy. Legal evaluation therefore needs to assess the quality of a model's reasons alongside the accuracy and consequences of its conclusions.

This approach can guide the development of interpretive evaluation in other domains. Start with a meaningful task, make its evidence and assumptions available, distinguish quality of support from degrees of plausibility, and test whether feedback leads to useful revision. Domain knowledge determines what counts as a relevant source or a sound connection. Each application requires its own rubric and validation.

Literature brings an important requirement into view: language models must account for how cultural expressions acquire meaning in particular uses. Meeting that requirement calls for literary scholarship to shape tasks and learning objectives, and for computational experiments to test which forms of evidence and feedback improve interpretation. Cultural context should shape the choice of training material, tasks, evaluation criteria, and feedback throughout model development. Progress can then be assessed through how well systems substantiate cultural claims, sustain plausible alternatives, and respond to informed criticism. Such evaluation provides a basis for developing culturally robust language technologies.

\clearpage
\begin{table*}[t]
\centering
\caption{Illustrative tasks for cultural referencing and reuse, with evidence and expert follow-up questions for each aspect of interpretation.}
\label{tab:tasks}
\small
\begin{tabularx}{\textwidth}{@{}p{0.17\textwidth}YYY@{}}
\toprule
\textbf{Task} & \textbf{Question for the model} & \textbf{Evidence to assess} & \textbf{Expert follow-up} \\
\midrule
Identify a reference & What earlier expression or text might be invoked? & A traceable source and a justified connection & Which features of the source support the proposed connection? \\
Interpret its use & What does the reference do in this passage? & Attention to speaker, narrative context, and relevant history & Where the expression recurs in another passage, how does its role differ? \\
Compare readings & What supports two plausible interpretations? & Accurate evidence and a fair account of disagreement & How does each reading accommodate evidence that supports another emphasis? \\
Revise after criticism & What changes if this evidence is mistaken or incomplete? & A revision that addresses the identified problem & Which claims change in response to the expert's textual or historical evidence? \\
\bottomrule
\end{tabularx}
\end{table*}
\clearpage

\section{Acknowledgments}
This work was supported by Independent Research Fund Denmark (DFF) under grant 5334-00088B, \emph{CuRe: Cultural Reasoning for Responsible Language Model Development}.

ChatGPT 6 (OpenAI) was used to assist with drafting, restructuring, and language revision. The authors reviewed and revised the resulting text and take responsibility for the content of the manuscript.

\section{Author contributions}
Daniel Hershcovich wrote the initial draft. All authors developed the argument, revised the manuscript, and approved its submission.

\section{Competing interests}
The authors declare no competing interests.

\section{Data availability}
No new data were generated or analyzed for this Perspective.

\bibliographystyle{abbrvnat}
\bibliography{references}

\end{document}